\documentclass{article}

\PassOptionsToPackage{numbers, compress}{natbib}

\usepackage[preprint]{neurips_2023}

\usepackage[utf8]{inputenc} 
\usepackage[T1]{fontenc}    
\usepackage{hyperref}       
\usepackage{url}            
\usepackage{booktabs}       
\usepackage{amsfonts}       
\usepackage{nicefrac}       
\usepackage{microtype}      
\usepackage{xcolor}         

\usepackage{graphicx}
\usepackage{subfigure}
\usepackage{textcomp}
\usepackage{multirow}
\usepackage{amsmath}
\usepackage{amssymb}
\usepackage{mathtools}
\usepackage{amsthm}
\usepackage{bbding}
\usepackage{bm}
\usepackage{arydshln}
\usepackage{wrapfig}

\title{TS-Diffusion: Generating Highly Complex \\ Time Series with Diffusion Models}

\author{%
  Yangming Li\thanks{Yangming Li is the first author of this work and other authors will be updated soon.} \\
  DAMTP, University of Cambridge\\
  \texttt{yl874@cam.ac.uk} \\
}

\begin{document}
\maketitle

\begin{abstract}
	
	While current generative models have achieved promising performances in time-series synthesis, they either make strong assumptions on the data format (e.g., regularities) or rely on pre-processing approaches (e.g., interpolations) to simplify the raw data. In this work, we consider a class of time series with three common bad properties, including sampling irregularities, missingness, and large feature-temporal dimensions, and introduce a general model, \textit{TS-Diffusion}, to process such complex time series. Our model consists of three parts under the framework of point process. The first part is an encoder of the neural ordinary differential equation (ODE) that converts time series into dense representations, with the jump technique to capture sampling irregularities and self-attention mechanism to handle missing values; The second component of TS-Diffusion is a diffusion model that learns from the representation of time series. These time-series representations can have a complex distribution because of their high dimensions; The third part is a decoder of another ODE that generates time series with irregularities and missing values given their representations. We have conducted extensive experiments on multiple time-series datasets, demonstrating that TS-Diffusion achieves excellent results on both conventional and complex time series and significantly outperforms previous baselines.
  
\end{abstract}

\section{Introduction}
	
	Machine learning (ML) has been widely used in time-series analysis~\citep{kim2003financial,zhao2017convolutional,ismail2019deep,paparrizos2015k}, but its application and development heavily rely on large datasets, which are not available in some sensitive domains (e.g., medicine and healthcare~\citep{zeger2006time}). One way to get around this problem is data synthesis~\citep{walonoski2018synthea,alaa2019attentive,yoon2020anonymization}, which learns generative models, such as GAN~\citep{NIPS2014_5ca3e9b1}, from sensitive data and applies them to generate fake datasets that are without privacy concern for sharing.
	
	Previous works~\cite{oord2016wavenet,esteban2017real,donahue2018adversarial} have already presented some effective models to time-series synthesis. For example, \citep{mogren2016c} proposed a GAN-based model, with LSTMs~\citep{hochreiter1997long} as the backbones of its generator and discriminator. While these models have shown promising results, they either make strong assumptions on the structure of time series or just pre-process the raw data. For instance, Time-GAN~\cite{NEURIPS2019_c9efe5f2} and Fourier Flows~\cite{alaa2021generative} supposed that the sampling intervals of time series are constant; While Latent ODE~\citep{NEURIPS2019-42a6845a} and GT-GAN~\citep{jeon2022gtgan} is applicable to irregular time series, they can't take data with missing values as inputs. Besides, the benchmark datasets to evaluate some models are not complex enough to expose their potential weaknesses. For example, Neural STPPs~\cite{chen2021neural} are only tested by sequence data of very low feature dimensions (i.e., $2$).

	We identify three prevalent undesirable characteristics found in real-world time series that undermine the effectiveness of current methodologies. These properties include: 1) \textbf{sampling irregularities}, where the time gaps between observations are not fixed; 2) \textbf{missingness}, which means the feature vector of every observation might contain missing values. For univariate time series, missingness is the same concept as irregularities, but this view is not correct for multivariate time series since it means to ignore the cross-dependencies among feature dimensions; 3) \textbf{large feature-temporal dimensionality}, where sizeable feature dimensions or large numbers of observations result in a huge overall dimensionality. We specifically define a class of time series with all these properties as \textit{highly complex}. An example is electronic health records (EHRs), which contain many incomplete medical tests at variable time points. Besides directly modeling these properties, another way to handle them is data preprocessing, such as imputations~\cite{gondara2017multiple,yoon2018gain} and interpolations~\citep{NEURIPS2020_4a5876b4,morrill2021neuralcontrolled}. However, its accuracy is not guaranteed and might cause information loss.
	
	In this work, we introduce a general model, \textit{TS-Diffusion}, that supports generating and learning from \textit{highly complex time series}, which is beyond the capabilities of current methods without resorting to data pre-processing. Under the framework of marked point process~\citep{daley2003introduction}, our model consists of three components. The first part is a  \textit{continuous-time encoder} constructed by the neural ordinary differential equation (ODE)~\citep{NEURIPS2018-69386f6b} with the jump technique~\cite{NEURIPS2019-59b1deff}, which converts multivariate time series of variable lengths and sampling irregularities into dense representations. Compared with other ODE encoders~\citep{NEURIPS2019-42a6845a,chen2021neural}, a notable difference is that our encoder is parameterized by multi-layer self-attentions~\citep{NIPS2017_3f5ee243} such that it's robust to missing values.

	The second component is a \textit{generative model}~\citep{sohl2015deep} that learns from the dense representations of time series. This task is challenging for conventional generative models (e.g., Normalizing Flow~\citep{dinh2017density} and VAE~\citep{kingma2013auto}) because the large dimensions of our data type result in a complex distribution of its representations. In recent studies~\citep{NEURIPS2021-49ad23d1,rombach2022high}, it has been consistently observed that diffusion models~\citep{sohl2015deep} exhibit exceptional performance in generating high-dimensional data (e.g., high-resolution images). Hence, we adopt DDPM~\citep{ho2020denoising}, a standard diffusion model, as the backbone of this component.
	
	The last part of TS-Diffusion is a \textit{continuous-time decoder} of another neural ODE. which is initialized by dense representations from the generative model and synthesizes irregular time series in a manner of inhomogeneous Poisson process~\cite{palm1943intensitatsschwankungen,NEURIPS2019-42a6845a}. For missingness, we properly define the \textit{observation probability} so that its marginalization over missing values is computationally feasible. With this tool, our model can learn from incomplete time series and generate such data at test time. For some time series, the correlation between observations is very strong. Therefore, we apply attention mechanism~\citep{bahdanau2014neural} to condition decoder states on formerly generated data.

	We have conducted extensive experiments to verify the effectiveness of TS-Diffusion. Firstly, we test our model on an EHR dataset, WARDS~\citep{alaa2017personalized}. The results indicate that our model significantly outperforms previous approaches in terms of log-likelihood and other automatic metrics (e.g., precision) related to data synthesis~\citep{sajjadi2018assessing}. Then, we show that our model also marginally outperforms the baselines for evaluations on conventional time-series datasets, such as Earthquakes~\citep{chen2021neural}. Lastly, we study two concepts about medical time series, informative sampling and treatment durations, and show that our model performs well in terms of them.

\section{Related Work}

	Our work is related to three research areas, including generative models, Neural ODEs, and point processes. In this section, we clarify our differences from previous works. We also diagram Table~\ref{tab:comparisons} to compare our proposed TS-Diffusion with key baselines in details.

	\paragraph{Classical Generative Models.} Many works~\citep{esteban2017real,alaa2019attentive,alaa2021generative,jeon2022gtgan} resort to classical generative models (e.g., VAE) to synthesize time series. For example, \citep{NEURIPS2019_c9efe5f2} presented TimeGAN, a variant of GAN that captured the autoregressive dependencies of time-series data. 
	These models mostly are inapplicable to highly complex time series without data pre-processing (e.g., imputations) because they make strong assumptions on the structure of time series. Notably, GT-GAN~\citep{jeon2022gtgan} is capable of processing both regular and irregular time series, but it can not handle missing values and its backbone, (e.g., GAN), underperforms diffusion models in data synthesis.

	\paragraph{Diffusion Models.}  As a new type of generative models, diffusion models have attracted much attention from the recent literature for their excellent performances ~\citep{ho2020denoising,nichol2021improved,saharia2022photorealistic}. For example, \citep{NEURIPS2021-49ad23d1} synthesized images of better quality than GAN with a standard diffusion model and \citep{rombach2022high} learns a separate VAE to embed images into low-dimensional representations for training diffusion models. Among these models, TabDDPM is the one that generates \textit{static} tabular data~\citep{kotelnikov2022tabddpm}. Although time series can be represented in a tabular format, their method does not consider any temporal dependency, and hence is not directly applicable to time series generation. 

	\paragraph{Neural ODEs.} While common sequential models (e.g., LSTM) suppose that input data are discrete and arrive at a fixed time interval, neural ODEs, which parameterize the derivatives of hidden layers with neural networks, are a continuous-time model of sequence data. For example, \citep{NEURIPS2020_4a5876b4} introduced neural CDEs that incorporate external observations into continuous hidden states of neural ODEs, and \citep{morrill2021neuralcontrolled} designed a novel interpolation scheme to extend them to online predictions. 
	While these models are naturally robust to sampling irregularities, they can't handle missing values and are commonly based on VAE, a relatively weak generative model.

	\begin{table*}
		\centering
		\small
		\setlength{\tabcolsep}{0.90mm}
		\begin{tabular}{c|cccc}
			\hline
			
			Model & Backbone & Irregularities & Missingness (Train) & Missingness (Eval.) \\
			\hline
			
			RMTPP~\citep{du2016recurrent} & Hawkes Process & \Checkmark & \textcolor{lightgray}{\XSolidBrush} & \textcolor{lightgray}{\XSolidBrush} \\
			
			HPM~\citep{Shelton_Qin_Shetty_2018} & Hawkes Process & \Checkmark & \Checkmark & \Checkmark \\
			
			Latent ODE~\cite{NEURIPS2019-42a6845a} & VAE & \Checkmark  & \textcolor{lightgray}{\XSolidBrush} & \textcolor{lightgray}{\XSolidBrush} \\
			
			Neural CDE~\citep{NEURIPS2020_4a5876b4} & VAE & \Checkmark & \Checkmark & \textcolor{lightgray}{\XSolidBrush} \\
			
			Fourier Flows~\citep{alaa2021generative} & Normalizing Flow & \textcolor{lightgray}{\XSolidBrush} & \textcolor{lightgray}{\XSolidBrush} & \textcolor{lightgray}{\XSolidBrush}\\
			
			Neural STPP~\citep{chen2021neural} & Normalizing Flow & \Checkmark & \textcolor{lightgray}{\XSolidBrush} & \textcolor{lightgray}{\XSolidBrush}\\
			
			TimeGAN~\citep{NEURIPS2019_c9efe5f2} & GAN & \textcolor{lightgray}{\XSolidBrush} & \textcolor{lightgray}{\XSolidBrush} & \textcolor{lightgray}{\XSolidBrush}  \\
			
			GT-GAN~\citep{jeon2022gtgan} & GAN & \Checkmark & \textcolor{lightgray}{\XSolidBrush}  & \textcolor{lightgray}{\XSolidBrush} \\
			
			\hline
			TS-Diffusion & Diffusion Model & \Checkmark & \Checkmark & \Checkmark \\
			\hline
			
		\end{tabular}
		
		\caption{Detailed comparisons between our model (i.e., TS-Diffusion) and key baselines. The last two columns respectively indicate whether a model is capable of learning from incomplete time series and generating such data at evaluation time. For the second column, note that diffusion models have shown amazing performances in synthesizing high-dimensional data~\citep{rombach2022high}.}
		\label{tab:comparisons}
	\end{table*}

	\paragraph{Point Processes.} Point processes are a standard paradigm to model a sequence of events that are drawn from a specific distribution with stochastic time intervals (i.e., sampling irregularity). Famous models include the Poisson process~\citep{kingman1992poisson} and the Hawkes process~\citep{mei2017neural}.
	The drawback of these models is that they generally have strong assumptions on the latent distributions of time series, leading to their failures in fitting real-world data. Following recent advances~\citep{NEURIPS2019-42a6845a,chen2021neural}, we adopt an inhomogeneous Poisson process~\cite{palm1943intensitatsschwankungen} with an intensity function that is conditional on all prior events.

\section{Model: TS-Diffusion}

	In this section, we first formulate the task of synthesizing \textit{highly complex time series} under the framework of point process, and then introduce our model, \textit{TS-Diffusion}.

\subsection{Task Definition}

	Unlike classical time series $\mathbf{X} = \{\mathbf{x}_1, \mathbf{x}_2, \cdots, \mathbf{x}_N\}$, where $N$ data points arrive with a fixed time gap, observed events in our data type (e.g., EHR) are recorded at irregular time intervals, which we accordingly denote as $\mathbf{E} = \{(\mathbf{x}_1, t_1), (\mathbf{x}_2, t_2), \dots, (\mathbf{x}_N, t_N)\}$. Specifically, every event $\mathbf{x}_i = [ x_{i,1}, x_{i,2}, \cdots, x_{i, D} ]^T$ is a $D$-dimensional feature vector that may contain missing values, and $t_i$ represents its occurrence time. Note that complex time series have high feature-temporal dimensions, which means at least one of event number $N$ and feature dimension $D$ is very large. Let $T^{\mathrm{max}}$ denote upper bound of time $t$, i.e., $t_i \in [0, T^{\mathrm{max}}], \forall i \le N$.

	The essence of generative models is to learn the potential data distribution $p(\mathbf{E})$. Through the concept of \textit{marked point process}~\citep{daley2003introduction}, its log-likelihood can be computed as
\begin{equation}
	\label{eq:loss}
	\log p(\mathbf{E}) = \sum_{i} \log \lambda(\mathbf{x}_i, t_i )  - \int \int \lambda(\mathbf{x}, t ) d\mathbf{x}dt,
\end{equation}
where $\lambda$ is the \textit{intensity function}. As a rough explanation to intensity $\lambda$, we can regard term $\lambda(\mathbf{x}, t) d\mathbf{x}dt$ as an approximation to conditional probability $P(\mathbf{x}' \in [\mathbf{x}, \mathbf{x} + d\mathbf{x}], t' \in [t, t + dt])$. The design of intensity $\lambda$ is central to the generative model~\citep{ozaki1979maximum,NIPS2017-6463c884,pmlr-v70-alaa17a,pmlr-v108-qian20a}.

\subsection{Overall Model}

	\begin{figure*}
		\centering
		\includegraphics[width=0.9\textwidth]{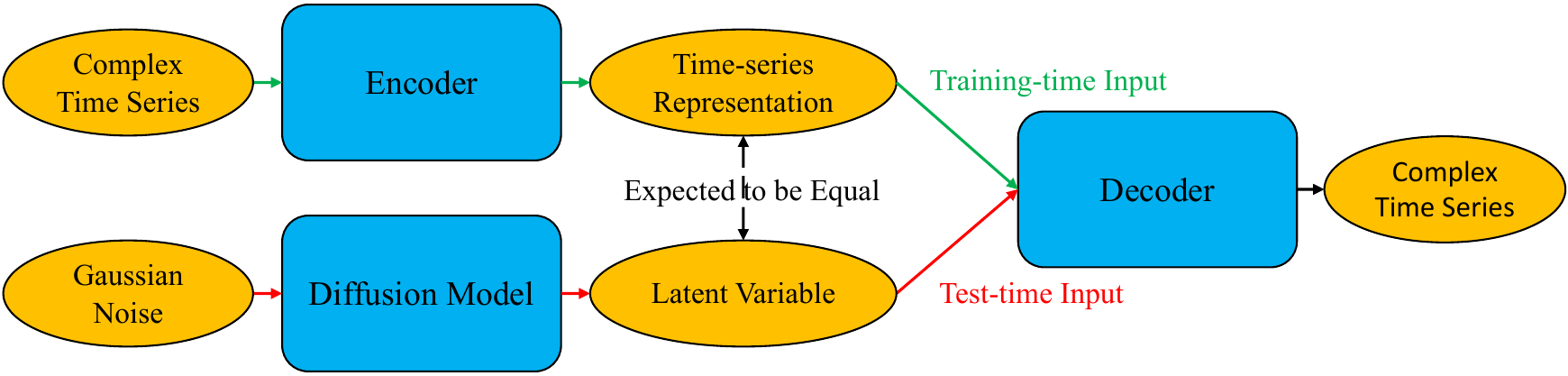}
		
		\caption{An overview of our model, TS-Diffusion. Complex time series means that the data are with three properties: irregularities, missingness, and large feature-temporal dimensions. Our model are constructed by three blocks: the encoder, the diffusion model, and the decoder. We specifically design them to handle these bad properties.}
		\label{fig:model demo}
	\end{figure*}

	As shown in Fig.~(\ref{fig:model demo}), our framework consists of three parts: 1) an encoder to cast time series $\mathbf{E}$ into a fixed-dimensional representation $\mathbf{s}$ that captures missing values and sampling irregularity; 2) a generative model that learns the distribution of time-series representation $\mathbf{s}$.
	We consider diffusion models for implementation, which perform amazingly well in generating high-dimensional data~\citep{rombach2022high}.
	3) a decoder that renders representation $\mathbf{s}$ into complex time series $\widehat{\mathbf{E}}$.

\subsubsection{Factorization of the Intensity Function}

	For simplifying the computation of Eq.~(\ref{eq:loss}), we apply a conclusion from \citep{daley2003introduction} to factorize intensity $\lambda(t, \mathbf{x})$ by the chain rule: $\lambda(\mathbf{x}, t) =  \lambda(t) p(\mathbf{x} \mid t)$, where $\lambda(t)$ is the intensity of purely temporal process $\mathbf{T} = [t_1, t_2, \dots, t_N]$ and $p(\mathbf{x} \mid t)$ is the \textit{observation probability} of event $\mathbf{x}$ at time $t$. For a clear formulation, we call $\lambda(\mathbf{x}, t)$ the marked intensity and $\lambda(t)$ the occurrence intensity. Through the above simplification, log-likelihood $\log p(\mathbf{E})$ can be reformulated as
	\begin{equation}
		\label{eq:simplified marked intensity}
		\log p(\mathbf{E}) = \sum_i \log \lambda(t_i)+ \sum_i \log p(\mathbf{x}_i \mid t_i) - \int \lambda(t) dt,
	\end{equation}
	which reduces the integral over $\mathbf{x}$.

\subsubsection{Encoder: Self-attentive Jump  ODE}
\label{sec:encoder}

	Complex time series are challenging for mainstream sequential models, e.g., LSTM, due to their missingness and irregularity. The encoder of our model applies self-attention~\citep{NEURIPS2019-59b1deff} and the jump technique in neural ODE~\citep{NEURIPS2019-59b1deff},  to respectively solve the two problems.

	\textbf{Self-attention Mechanism for Missingness.} For every observation $\mathbf{x}_i$, we first respectively embed its all $D$ dimensions and value $x_{i,j}$ into vectors $\mathbf{y}_{i,j}, 1 \le j \le D$ and $\mathbf{z}_{i,j}$:
	\begin{equation}
		\{ \mathbf{y}_{i,1}, \mathbf{y}_{i,2}, \cdots, \mathbf{y}_{i,D} \} = \{ \mathbf{u}_1, \mathbf{u}_2, \cdots, \mathbf{u}_D \}, \ \ \ \mathbf{z}_{i,j} = x_{i,j} \cdot \mathbf{1}, 1 \le j \le D,
	\end{equation}
	where $\mathbf{u}_j, 1 \le j \le D$ are learnable vectors and $\mathbf{1}$ denotes a vector with all values being $1$. Then, inspired by \citep{chen-etal-2017-enhanced}, we combine the two groups of representations as
	\begin{equation}
		\mathbf{e}_{i,j}^1  = \mathbf{y}_{i,j} \oplus \mathbf{z}_{i,j} \oplus (\mathbf{y}_{i,j} - \mathbf{z}_{i,j}) \oplus (\mathbf{y}_{i,j} \odot \mathbf{z}_{i,j}), \ \ \ \mathbf{e}_{i,j}^2 = \mathbf{W}^2_e\tanh(\mathbf{W}^1_e\mathbf{e}^1_{i,j}),
	\end{equation}
	where $\mathbf{W}^1_e, \mathbf{W}^2_e$ are learnable matrices, $\oplus$ means column-wise vector concatenation, and $\odot$ is element-wise vector product. Note that some values of variable $\mathbf{x}_i$ may be missing. We fill them with $0$ to benefit from parallel matrix computations on GPUs and adopt a mask vector $\mathbf{m}_i \in \{0, 1\}^D$ to indicate whether value $x_{i,j}$ is missing (i.e., $m_{i,j} = 0$) or not (i.e., $m_{i,j} = 1$). This vector will be used by self-attentions to avoid involving missing values into computations.

	Lastly, we apply self-attention mechanism to compute the representation of variable $\mathbf{x}_i$ with all its dimension representations $\mathbf{e}^2_{i,j}, 1 \le j \le D$:
	\begin{equation}
			\{\mathbf{q}_{i,j}, \mathbf{k}_{i,j}, \mathbf{v}_{i,j}\} = \{\mathbf{Q}\mathbf{e}^2_{i,j}, \mathbf{K} \mathbf{e}^2_{i,j}, \mathbf{V}\mathbf{e}^2_{i,j}\}, \ \ \ \widetilde{\mathbf{x}}_i = \sum_{j} \Big( \frac{ m_{i,j} * \exp( \mathbf{q}_{i,j}^T \mathbf{k}_{i,j})}{\sum_{j'} m_{i,j'} * \exp(\mathbf{q}_{i,j'}^T \mathbf{k}_{i,j'})} \Big) \mathbf{v}_{i,j},
	\end{equation}
	where matrices $\mathbf{Q}, \mathbf{K}, \mathbf{V}$ are all trainable parameters. Following common practices~\citep{NIPS2017_3f5ee243,devlin-etal-2019-bert,dosovitskiy2021an}, we will carry out this process many times to form multiple layers of self-attention and apply residual learning~\citep{he2016deep} and batch normalization~\citep{ioffe2015batch} to all these layers. For notational convenience, we omit these details and regard vector $\widetilde{\mathbf{x}}_i$ as the output of the last layer.

	\textbf{Jump ODE for Irregularities.}	We also apply a neural ODE to model the latent continuity between adjacent events and the jump technique to capture their occurrence patterns:
	\begin{equation}
		\label{eq:encoder ode}
			\frac{d\mathbf{s}_t}{dt}  = f_s(\mathbf{s}_t, t), \forall t \notin \mathbf{T}, \ \ \
			\lim_{\epsilon \in \mathbb{R}^+, \epsilon \rightarrow 0} \mathbf{s}_{t - \epsilon} = g_s(\mathbf{s}_t, \widetilde{\mathbf{x}}_{i}, t), t = t_i, \forall t_i \in \mathbf{T},
	\end{equation}
	where $f_s$ and $g_s$ are respectively parameterized by multi-layer perceptrons (MLP) and LSTM. Because of the discontinuous jumps made by $g_s$, representation $\mathbf{s}_t$ is right-continuous with left limits. We set the encoder dynamics to run backward in time, i.e. starting at $t = T^{\mathrm{max}}$ and ends at $t = 0$ with state $\mathbf{h}_{T^{\mathrm{max}}}$ being initialized by a learnable vector. We set the dense representation of data $\mathbf{E}$ as ODE state $\mathbf{s}_{0}$. For brevity, we omit its subscript and denote the term as $\mathbf{s}$.
	Obtaining $\mathbf{s}$ amounts to solving an initial value problem (IVP). We follow \citep{NEURIPS2018-69386f6b} to use the sensitivity
	method to avoid the large memory overhead for  backward differentiation.

\subsubsection{Generative Model: DDPM}
\label{sec:generative}

	The second part of TS-Diffusion is to learn the  distribution of representation $\mathbf{s}$ for generating high-fidelity and diverse time series.
	We implement this part via diffusion models, a powerful generative model that has received much attention from recent works~\citep{NEURIPS2021-49ad23d1,nichol2021improved,rombach2022high}.

	A standard diffusion model, DDPM~\citep{ho2020denoising}, is composed of a forward and backward Markovian processes. In the first process (also called the diffusion process), noises are incrementally incorporated into a sample at a predefined number of iterations $L \gg 1$ such that its final distribution approximately turns into an isotropic Gaussian distribution:
	\begin{equation}
		\label{eq:diffusion}
		p(\mathbf{h}_{k} \mid \mathbf{h}_{k-1}) = \mathcal{N}(\mathbf{h}_k; \sqrt{1 - \beta_k}\mathbf{h}_{k-1}, \beta_k\mathbf{I}),\\
	\end{equation}
	where $\mathbf{h}_0 = \mathbf{s}$, $1 \le k \le L$ and $\mathbf{I}$ is an identical matrix. 
	The other process (also known as the reverse process) is essentially a sequence of conditional probabilities $p(\mathbf{h}_{k-1} \mid \mathbf{h}_{k}), L \ge k \ge 1$, which iteratively denoise a Gaussian noise $\mathbf{h}_L$ into a true sample $\mathbf{h}_0$ that is of the same distribution as time-series representation $\mathbf{s}$. However, directly estimating probability $p(\mathbf{h}_{k-1} \mid \mathbf{h}_{k})$ is computationally infeasible. Current practices mainly approximate it as a Gaussian distribution:
	\begin{equation}
		\label{eq:backward iter}
		q(\mathbf{h}_{k-1} \mid \mathbf{h}_k) = \mathcal{N}(\mathbf{h}_{k-1}; \bm{\mu}_k, \bm{\Sigma}_k),
	\end{equation}
	in which mean $\bm{\mu}_k$ and variance $\bm{\Sigma}_k$ are either model predictions or fixed as proper values. To reduce the variance of loss estimation, mean $\bm{\mu}_k$ is parameterized as
	\begin{equation}
		\label{eq:reverse mean}
		\bm{\mu}_k = \frac{1}{\sqrt{\alpha_k}} \Big( \mathbf{h}_k - \frac{\beta_k}{\sqrt{1 - \overline{\alpha}_k}}\bm{\epsilon}_q (\mathbf{h}_k, k) \Big),
	\end{equation}
	where $\alpha_k = 1 - \beta_k$, $\overline{\alpha}_k = \prod_{k'=1}^{k} \alpha_{k'}$ and $\bm{\epsilon}_q$ is a neural network to fit Gaussian noises, and variance $\bm{\Sigma}_k$ is fixed as $\beta_k \mathbf{I}$ or $\beta_k (1 - \overline{\alpha}_{k-1}) \ (1 - \overline{\alpha}_k)\mathbf{I}$, where $\overline{\alpha}_0 = 1$. 
	
	\paragraph{Roles in TS-Diffusion.}  The diffusion model learns from time-series representation $\mathbf{s}$ at training time, which is also fed into the decoder. Like VAE, we also apply a few diffusion steps with Eq.~(\ref{eq:diffusion}) to slightly add noises into representation $\mathbf{s}$.
	For evaluation, we first sample Gaussian noise $h_T \sim \mathcal{N}(\mathbf{0}, \mathbf{I})$ and iteratively denoise it into time-series representation $\mathbf{h}_0$ through the backward process, which is then conditioned by the decoder to generate complex time series.

\subsubsection{Decoder: ODE and Marked Point Process}
\label{sec:decoder}

	Our decoder is used to generate time series. Its input is either latent variable $\mathbf{h}_0$ from the diffusion model at evaluation time or time-series representation $\mathbf{s}$ from the encoder during training. In the following, we only use term $\mathbf{s}$ for notational convenience.

	We adopt another neural ODE as the backbone of this module, which defines how the latent state $\mathbf{o}_t$ is initialized by time-series representation $\mathbf{s}$ and evolves over time $t$. The dynamics of state $\mathbf{o}_t$ is conditioned on all its observable events $\mathbf{E}_{<t} = \{(\mathbf{x}_i, t_i) \mid t_i < t \}$ as
	\begin{equation}
		\label{eq:decoder ode}
			\mathbf{o}_{0} = f_o(\mathbf{s}), \ \ \
			\frac{d\mathbf{o}_{t}}{dt} = g_o(\mathbf{o}_t, t, \mathbf{E}_{<t}),
	\end{equation}
	in which $f_o$ is a simple MLP and $g_o$ is implemented by attention mechanism:
	\begin{equation}
		\label{eq:dec attention}
		\mathbf{a}_t = \sum_{t_i < t} \Big( \frac{\exp((\mathbf{o}_t')^T\mathbf{x}_i')}{\sum_{t_{j} < t} \exp((\mathbf{o}_t')^T\mathbf{x}_j')} \Big) \mathbf{x}_i', \ \ \ \frac{d\mathbf{o}_{t}}{dt} = \mathbf{W}^2_g\tanh(\mathbf{W}^1_g (\mathbf{o}_t' \oplus \mathbf{a}_t)),
	\end{equation}
	where $\mathbf{o}_t', \mathbf{x}_j'$ are time-sensitive representations:
	\begin{equation}
		\mathbf{o}_t' = \mathbf{W}^2_o\tanh(\mathbf{W}^1_o(\mathbf{o}_t \oplus (t \cdot \mathbf{1}))), \ \ \ \mathbf{x}_i'  = \mathbf{W}^2_x \tanh(\mathbf{W}^1_x(\widetilde{\mathbf{x}}_i \oplus (t_i \cdot \mathbf{1}))),
	\end{equation} 
	where $\mathbf{W}^i_k, i \in \{1, 2\}, k \in \{o, x, g\}$ are learnable matrices. At any time $t$, we predict the factorized parts of $\lambda(t, \mathbf{x})$, i.e., occurrence intensity $\lambda(t)$ and observation probability $p(\mathbf{x} \mid t)$, as
	\begin{equation}
		\label{eq:marked intensity}
			\lambda(t) = \log(1 + \exp(\mathbf{w}^T_{\lambda}\mathbf{o}_t)), \ \ \
			p(\mathbf{x} \mid t) \propto \exp( -\frac{1}{2}|| \mathbf{x} - \mathbf{W}_p^T\mathbf{o}_t||^2_2  ),
	\end{equation}
	where vector $\mathbf{w}_{\lambda}$ and matrix $\mathbf{W}_p$ are learnable parameters, and operation $||\cdot||_2$ is $\mathbb{L}^2$ norm. Note that $p(\mathbf{x} \mid t)$ is essentially a multivariate Gaussian $\mathcal{N}(\mathbf{x}; \mathbf{W}_p^T\mathbf{o}_t, \mathbf{I})$, so we pre-normalize the variance of each dimension of observation $\mathbf{x}$ to a unit scale.

	\paragraph{Handling Missingness.} We adopt a marginalization approach~\citep{tsuboi2008training} to train our model with incomplete time series. Since the multivariate Gaussian has a closed-form solution to its marginalization over any part of its variables, we can re-formulate $p(\mathbf{x} \mid t)$ in Eq.~(\ref{eq:marked intensity}) as
	\begin{equation}\nonumber
		p(\mathbf{x} \mid t, \mathbf{m}) = \frac{1}{(\sqrt{2\pi})^{\sum_j m_j}} \exp\Big(-\frac{||( \mathbf{x} - \mathbf{W}_p^T\mathbf{o}_t) \odot \mathbf{m} ||_2^2}{2}\Big),
	\end{equation}
	where mask vector $\mathbf{m} = [m_1, m_2, \cdots, m_D]^T$ is formerly defined in Sec.~\ref{sec:encoder} to indicate the positions of missing values in observation $\mathbf{x}$. Besides robust training, we parameterize a learnable multivariate Bernoulli distribution of the missing values in training data:
	\begin{equation}
		\label{eq:missingness def}
		\widehat{\mathbf{m}} = \mathrm{sigmoid}(\mathbf{W}^2_m \tanh(\mathbf{W}^1_m ((t \cdot \mathbf{1}) \oplus \mathbf{o}_t))).
	\end{equation}
	At evaluation time, we can simulate whether each value $x_j, 1 \le j \le D$ of observation $\mathbf{x}$ is missing through sampling from Bernoulli distribution $\mathcal{B}(\widehat{m}_j)$.
	
	\paragraph{Variable Time Horizon.} 
	Previous works~\citep{NEURIPS2019-42a6845a,chen2021neural} generally fix $T^{\mathrm{max}}$ as a global constant, i.e., the maximum end time $t_{N}$ for the whole training set. However, this assumption is too strong. For example, we observe that patients recorded in EHRs have highly different treatment duration. To permit inconstant horizon $T^{\mathrm{max}}$, we introduce a special module to capture its variability:
	\begin{equation}
		\mu_t = \mathbf{W}^2_t\tanh(\mathbf{W}^1_t \mathbf{s}), \ \ \ \widehat{T}^{\mathrm{max}} \sim \mathcal{N}(\mu_t,  \sigma_t ),
		\label{eq:time}
	\end{equation}
	where matrices $\mathbf{W}^1_t, \mathbf{W}^2_t$ are learnable parameters and $\sigma_t$ denotes the empirical variance of end time $t_N$ estimated from the whole training set.

\subsubsection{Loss Functions for Optimization}
\label{sec:losses}

	The optimization of our model is to minimize a hybrid loss that contains four parts. The first of them is negative log-likelihood of an EHR $\mathcal{L}_1 = -\log p(\mathbf{E})$, which is derived via Eq.~(\ref{eq:simplified marked intensity}) with occurrence intensity $\lambda(t)$ and observation probability $p(\mathbf{x} \mid t)$ computed by Eq.~(\ref{eq:marked intensity}). To efficiently estimate integral $\int \lambda(t) dt$ in Eq.~(\ref{eq:simplified marked intensity}), we jointly solve it and the ODE of state $\mathbf{o}_t$, i.e., Eq.~(\ref{eq:decoder ode}), with intensity $\lambda(t)$ defined in Eq.~(\ref{eq:marked intensity}), by numerical integration.

	The second one is the expected log-likelihood of real sample $\mathbf{h}_0 = \mathbf{s}$ for training the diffusion model: $\mathcal{L}_2 = \mathbb{E}[-\log q(\mathbf{h}_0)]$. Because this term is computationally infeasible, we follow common practices~\citep{ho2020denoising} to adopt its lower bound, which can be further simplified as
	\begin{equation}
		\label{eq:reparameterized loss}
		\mathcal{L}_2' = \mathbb{E}_{t, \bm{\epsilon}} \Big[ || \bm{\epsilon} - \bm{\epsilon}_q(\sqrt{\overline{\alpha}_k} \mathbf{h}_0 + \sqrt{1 - \overline{\alpha}_k} \bm{\epsilon}, t) ||_2^2 \Big],
	\end{equation}
	where step $t$ and noise $\bm{\epsilon}$ are respectively sampled from uniform distribution $\mathcal{U}\{1, L\}$ and standard normal distribution $\mathcal{N}(\mathbf{0}, \mathbf{I})$. 

	The third part is to optimize the module used to sample upper time bound $\widehat{T}^{\mathrm{max}}$. Based on maximum likelihood estimation (MLE), we incur the loss as $\mathcal{L}_3 = ( t_N * (1 + \delta) -  \mu_t )^2$,
	where $\delta > 0$ is a predefined small value, which guarantees $\widehat{T}^{\mathrm{max}} > t_N$. This setup is very intuitive. For example, the discharge of a patient from the hospital is usually a little later than the last medical test. The last part is to train the model that simulates missingness at test time. Based on its definition in Eq.~(\ref{eq:missingness def}), we apply cross entropy to define the loss function as
	\begin{equation}
		\mathcal{L}_4 = \sum_{i=1}^{N}\sum_{j=1}^{D} \Big( m_{i,j} \ln(\widehat{m}_{i,j}) + (1 - m_{i,j}) \ln (1 - \widehat{m}_{i,j}) \Big),
	\end{equation}
	where $m_{i,j}$ indicates whether the $j$-th value of observation $x_i$ is missing.
	With all above results, our hybrid loss is shaped as $\mathcal{L} = \lambda_1 \mathcal{L}_1 + \lambda_2 \mathcal{L}_2' + \lambda_3 \mathcal{L}_3 +  \lambda_4 \mathcal{L}_4$,
	where positive weights $\lambda_1, \lambda_2, \lambda_3$ are predefined to balance different losses.

\subsubsection{Sampling for Data Synthesis}

	Data synthesis is the process of sampling from a generative model (i.e., learned data distribution). For TS-Diffusion, it contains three stages: 1) the backward process of the diffusion model, which first draws a sample from Gaussian distribution $\mathcal{N}(\mathbf{0}, \mathbf{I})$ and then iteratively denoises it into latent variable $\mathbf{s}$; 2) we sample the maximum time horizon $\widehat{T}^{\mathrm{max}}$ using Eq.~(\ref{eq:time}); 3) the decoder is conditioned on input sample $\mathbf{s}$ to compute marked intensity $\lambda(\mathbf{x}, t)$ over the time horizon and the standard thinning algorithm~\citep{lewis1979simulation} is applied to generate a marked point process from it. If it's necessary, we can also predict the mask vector $\widehat{\mathbf{m}}$ to simulate missingness in the generated data.

\section{Experiments}

	\begin{table*}
		\centering
		\small
		\setlength{\tabcolsep}{1.1mm}
		{\begin{tabular}{c|c|cccc|c}
				\hline
				
				\multicolumn{2}{c|}{Method} & RMTPP~\citep{du2016recurrent} & Neural CDE~\citep{NEURIPS2020_4a5876b4} & Latent ODE~\cite{NEURIPS2019-42a6845a}  & Neural STPP~\citep{chen2021neural} & TS-Diffusion  \\ 
				
				\hline
				\multirow{2}{*}{WARDS} & Temporal & $0.873$ & $1.153$ & $1.082$ & $1.231$ & $\mathbf{1.819}$ \\
				& Feature & $-35.628$ & $-31.749$ & $-33.287$ & $-32.692$ & $\mathbf{-28.835}$  \\
				
				\hline
				
		\end{tabular}}
		\caption{Log-likelihood scores per observation of our model and baselines on the test set of WARDS. All scores here are averaged over $5$ runs with standard deviations smaller than $0.05$.}
		\label{tab:likelihood on wards}
	\end{table*}

	To verify the effectiveness of our model, we first compare it with key baselines on multiple representative datasets of time series. Then, we perform some special studies (e.g., informative sampling) and show that TS-Diffusion generates complex time series of high quality.

\subsection{Settings}

	Three benchmark datasets are used in our experiments. One is WARDS~\citep{alaa2017personalized}, a medical dataset consisting of \textit{highly complex time series}.
	The measurement result is a $37$-dimensional vector and the maximum sequence length is close to $1000$, offering a good example of high feature-temporal dimensionality. In addition, as a result of irregular clinical measurements, the intervals between two measurement times are not fixed and every measurement vector may have missing values. The other two are conventional time-series datasets introduced by a strong baseline, Neural STPP~\citep{chen2021neural}, Earthquakes and COVID-19 Cases. 
	Both datasets contain irregular time series but they don't have missing values and their feature dimensions are $2$.

	We use almost the same model configurations for all experiments. The number of self-attention layers in the encoder is $3$. The hidden dimensions of the encoder and decoder are both $128$. The number of forward or backward iterations of the diffusion model is set as $1000$. Loss weights $\lambda_1, \lambda_2, \lambda_3, \lambda_4$ are respectively set as $0.4, 0.4, 0.1, 0.1$. We adopt Adam algorithm~\citep{kingma:adam} with the default hyper-parameter setting to optimize our model and set dropout ratio as $0.1$ to avoid overfitting. Our models are trained on 1 NVIDIA Tesla V100 GPU within 3 days.

\subsection{Evaluations on Highly Complex Time Series}

	We use the WARDS dataset to conduct the evaluations. Because it has missing values that are beyond the reach of some baselines, we apply data imputations~\citep{yoon2018gain} to pre-process the training data for them. Table \ref{tab:likelihood on wards} compares our model with key baselines by the log-likelihood score computed by Eq.~(\ref{eq:reparameterized loss}). 
	\setlength{\intextsep}{0pt}
	\begin{wrapfigure}{t}{0.35\textwidth}
		\centering
		\includegraphics[width=0.35\textwidth]{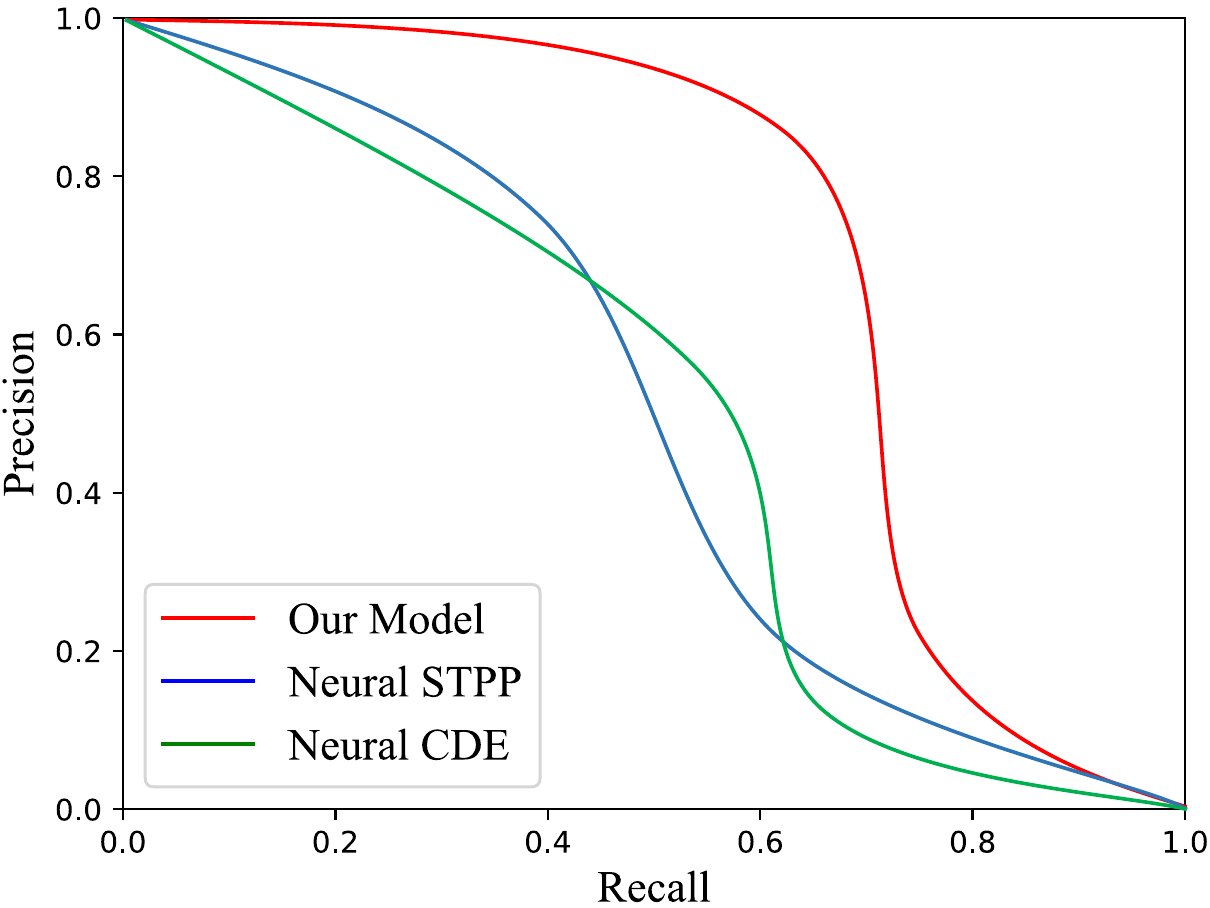}
		
		\caption{Precision-recall curves of some models on WARDS.}
		\label{fig:prd curves}
	\end{wrapfigure}
	We also divide its mean into two part: temporal score $\frac{1}{N}(\sum_i \log \lambda(t_i) - \int \lambda(t) dt)$ and feature score $\frac{1}{N}\sum_i \log p(\mathbf{x}_i \mid t_i)$. All baselines, which have official open-source codes, and our model learns from the training set and we estimate their scores on the test set. 
	From the table, we can see that our model notably outperforms previous baselines, with $47.76\%$ improvements on temporal log-likelihood and $2.91$ point increases on the feature score. These results suggest our model does better than baselines on complex time series.

	As depicted in Fig.~\ref{fig:prd curves}, we evaluate our model and two baselines, Neural STPP and Neural CDE, on the WARDS dataset with precision and recall metrics introduced by \citep{sajjadi2018assessing}. Roughly speaking, precision measures the quality of generated samples and recall implies their diversity. 
	We can draw two conclusions from the curves: 1) our curve locates far higher than those of the baselines, indicating our model synthesizes samples of higher fidelity and diversity; 2) the curve of Neural STPP is close to that of Neural CDE, which matches the results shown in Table~\ref{tab:likelihood on wards}. This study further confirms that our framework is much more effective than the baselines for applications to complex time series.

		\begin{table*}
		\centering
		\small
		\setlength{\tabcolsep}{1.5mm}
		{\begin{tabular}{c|cc|cc}
				\hline
				
				\multirow{2}{*}{Model} & \multicolumn{2}{c|}{Earthquakes} & \multicolumn{2}{c}{COVID-19} \\
				\cline{2-5}
				
				& Temporal & Feature & Temporal & Feature \\
				\hline
				
				Neural Hawkes Process~\citep{mei2017neural} & $-0.198_{\pm 0.001}$  & $-$ & $2.229_{\pm 0.013} $ & $-$ \\
				
				Conditional KDE~\citep{reinhart2018review} & $-$ & $-2.259_{\pm 0.001}$ & $-$ & $-2.583_{\pm 0.000} $ \\
				
				Time-varying CNF~\citep{NEURIPS2018-69386f6b} & $-$ & $-1.459_{\pm 0.016}$ & $-$ & $-2.002_{\pm 0.002}$ \\
				
				Neural CDE~\citep{NEURIPS2020_4a5876b4} &  $0.178_{\pm 0.002}$ & $-1.155_{\pm 0.031}$ & $2.164_{\pm 0.005}$ & $-2.017_{\pm 0.001}$ \\
				
				Neural STPP w/ Jump CNF~\citep{chen2021neural} & $0.166_{\pm 0.001}$ & $-1.007_{\pm 0.050}$ & $2.242_{\pm 0.002}$ & $-1.904_{\pm 0.004}$ \\
				
				Neural STPP w/ Attentive CNF~\citep{chen2021neural} & $0.204_{\pm 0.001}$ & $-1.237_{\pm 0.075}$ & $2.258_{\pm 0.002}$ & $-1.864_{\pm 0.001}$ \\
				
				\hline
				TS-Diffusion & $\mathbf{0.219}_{\pm 0.002}$ & $-1.365_{\pm 0.028}$ & $2.375_{\pm 0.004}$ & $-1.756_{\pm 0.002}$ \\
				
				TS-Diffusion w/ Vanilla CNF & $0.213_{\pm 0.001}$ & $\mathbf{-1.003}_{\pm 0.032}$ & $\mathbf{2.388}_{\pm 0.005}$ & $\mathbf{-1.732}_{\pm 0.001}$ \\
				\hline
				
		\end{tabular}}
		\caption{Log-likelihood scores per observation of our models and baselines on the test sets of two conventional datasets. Every score is paired with a standard deviation estimated over $3$ runs.}
		\label{tab:likelihood on conventional data}
	\end{table*}

	\begin{table*}
		\centering
		\small
		\setlength{\tabcolsep}{0.6mm}
		{\begin{tabular}{c|ccccc|c}
				\hline
				
				Model & Latent ODE~\cite{NEURIPS2019-42a6845a} & Neural CDE~\citep{NEURIPS2020_4a5876b4} & Neural STPP~\citep{chen2021neural} & GT-GAN~\citep{jeon2022gtgan} & Oracle & TS-Diffusion \\
				
				\hline
				TFC $\rho_{t, x}$ ($\%$) & $1.57$  & $2.95$ & $3.15$ & $2.83$ & $3.92$ & $\mathbf{3.61}$ \\
				
				\hline
		\end{tabular}}
		\caption{TFC scores of our model and key baselines on the WARDS dataset. The oracle score is computed from the test set, indicating the upper bound that a model can reach.}
		\label{tab:informative sampling}
	\end{table*}

\subsection{Evaluations on Conventional Time series}

	We also have performed evaluations on two conventional time-series datasets: Earthquakes and COVID-19 Cases. Table~\ref{tab:likelihood on conventional data} diagrams the performances of our models and the baselines. While the scores of Neural CDE are from our evaluation, we follow the results of all other baselines reported by \citep{chen2021neural}. Besides, since the feature distribution of Earthquakes is very complex and both datasets don't have missing values, we follow Neural STPP to replace the standard Gaussian to compute observation probability $p(\mathbf{x} \mid t)$ in Eq.~(\ref{eq:marked intensity}) with continuous normalzing flow (CNF)~\citep{NEURIPS2018-69386f6b}.
	
	From the table, we can see that either our model or it with CNF achieves the best performances on both datasets. For example, in terms of temporal scores, our models outperform the best practice, Neural STPP, by $4.41\%$ on Earthquakes and $5.76\%$ on COVID-19 Cases. These results indicate that our framework is applicable to both complex and conventional time series.

	\begin{figure*}
		\centering
		\includegraphics[width=0.72\textwidth]{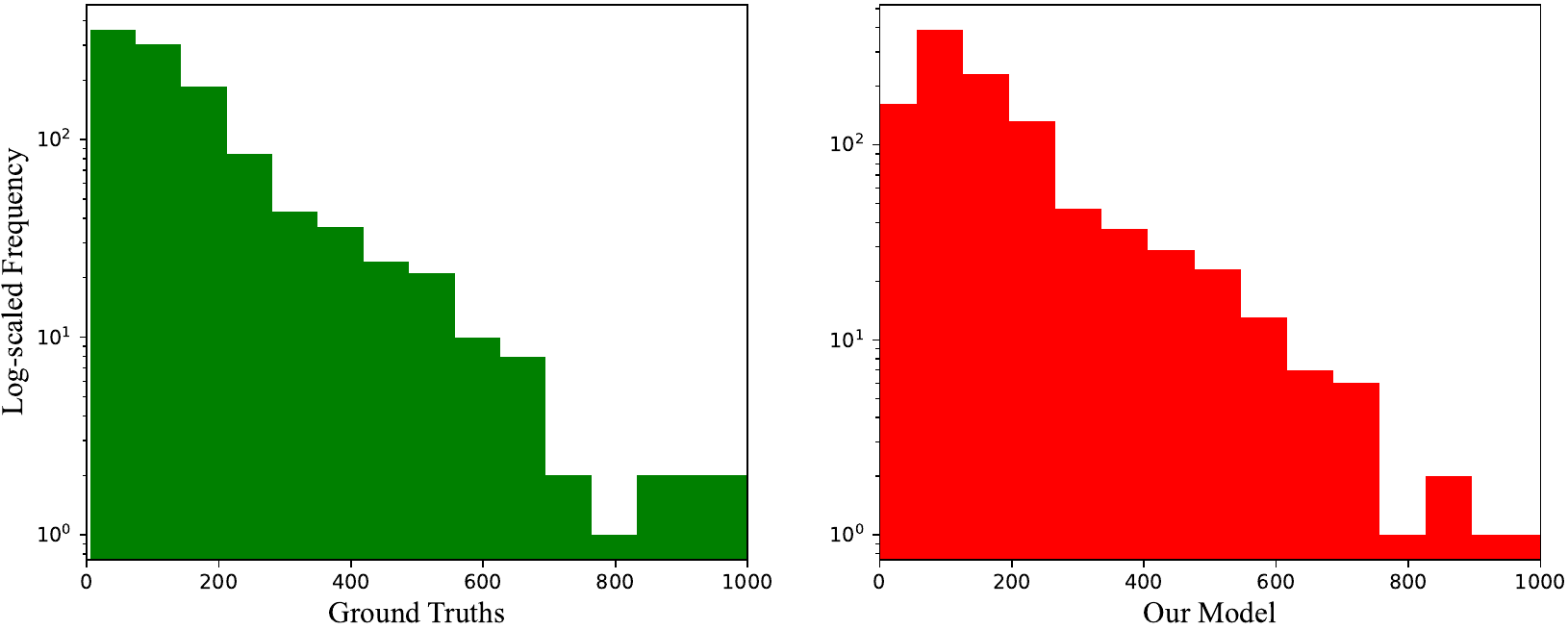}
		
		\caption{Empirical distributions of treatment durations. The left figure is from test set of the WARDS dataset and the right one is from synthetic data generated by our model.}
		\label{fig:durations}
	\end{figure*}

\subsection{Study on Informative Sampling}

	We are interested in whether our model is capable of capturing the rich information of relevance between sampling times and features. This type of relevance is also referred to as informative sampling in the field of medical time series~\citep{pmlr-v70-alaa17a}. For example, a patient in an abnormal clinical state will frequently receive medical diagnoses. We introduce a metric, temporal-feature correlation (TFC), that partially measures how a generative model performs in terms of informative sampling. To compute the metric, we first let the generative model synthesize a bunch of time series $\widehat{\mathbf{E}}$ and concatenate them into a large set $\widehat{\mathcal{S}}$. Then, we compute Pearson's correlation coefficient $\rho_{t, j}$ betwen sampling time $t$ and every dimension $j, 1 \le j \le D$ of feature $\mathbf{x}$ based on all pairs $(\widehat{\mathbf{x}}, \widehat{t})$ from set $\widehat{\mathcal{S}}$. Finally, we get TFC score $\rho_{t, x} = \frac{1}{D} \sum_j | \rho_{t, j} |$.

	\begin{table*}
		\centering
		\small
		\setlength{\tabcolsep}{0.6mm}
		{\begin{tabular}{c|cc}
				\hline
				
				Model & Temporal (WARDS) & Feature (WARDS) \\
				
				\hline
				TS-Diffusion & $\mathbf{1.819}$ & $\mathbf{-28.835}$ \\
				
				\hdashline
				TS-Diffusion w/o Jump ODE Encoder, w/ LSTM & $1.568$ & $-33.549$ \\
				
				TS-Diffusion w/o Decoder Attention & $1.731$ & $-30.192$ \\
				
				TS-Diffusion w/o Variable Time Horizon & $1.653$ & $-29.083$ \\
				
				\hline
		\end{tabular}}
		\caption{Ablation experiments to verify the effectiveness of some parts of TS-Diffusion.}
		\label{tab:abalations}
	\end{table*}

	Table \ref{tab:informative sampling} shows the TFC scores of our model and the baselines on WARDS, a highly complex time-series dataset. Importantly, we also compute the upper bound of TFC by using the test set to construct set $\widehat{\mathcal{S}}$. From the table, we can see that our score is very close to the oracle with a gap of 0.31 percentage points and outnumbers those of baselines by at last $14.60\%$. Notably, we also involve GT-GAN into the comparison, with $0.78$ points lower than our performance. These results imply that our model performs well in terms of informative sampling. We attribute this to the attention mechanism used in our decoder.

\subsection{Study on Treatment Durations}

	Treatment duration $t_N - t_0$, which means the period length of a patient receiving medical tests, is also of our interest to evaluate the quality of synthetic data. In Sec.~\ref{sec:decoder}, we specifically set a module for our framework to sample end time $T^{\mathrm{max}} \approx (1 + \delta) * (t_N - t_0)$, while previous models mostly adopt fixed end time. From Fig.~\ref{fig:durations}, we can see that the duration distribution produced by our model is very close to that of the test set.

\subsection{Ablation Studies}

	Table \ref{tab:abalations} shows the effects of some components of our model, including Jump ODE defined in Eq.~(\ref{eq:encoder ode}) to capture sampling irregularities, decoder attention defined in Eq.~(\ref{eq:dec attention}) to model the dependencies between adjacent observations, and the setup of variable time horizon. The results show that all these components contribute to the great performances of TS-Diffusion.

\section{Conclusion}

	TS-Diffusion is a general model for generating \textit{highly complex time series} with irregularities, missingness, and the large feature-temporal dimensionality, which are aspects overlooked by previous works. Its excellent performances are strongly confirmed through extensive experiments.  
	We envision that TS-Diffusion can be applied to diverse application domains (such as healthcare and finance) to supply the practitioners with a powerful tool that can synthesize time series of a complex structure without resorting to data pre-processing.

\bibliography{neurips_2023}
\bibliographystyle{unsrt}

\end{document}